\documentclass[runningheads]{llncs}
\usepackage{graphicx}
\usepackage{subfigure}
\usepackage{tabularx}
\usepackage{amsmath}
\usepackage{amsfonts}
\usepackage{multirow}
\usepackage{hyperref}
\hypersetup{
    colorlinks=true,
    linkcolor=blue,
    filecolor=blue,      
    urlcolor=blue,
    citecolor=cyan,
}
\begin{document}
\title{Instance-based Vision Transformer for Subtyping of Papillary Renal Cell Carcinoma in Histopathological Image}
%
%

\author{Zeyu Gao\inst{1,2} \and
Bangyang Hong\inst{1,2} \and
Xianli Zhang\inst{1,2} \and
Yang Li\inst{1,2} \and
Chang Jia\inst{1,2} \and
Jialun Wu\inst{1,2} \and
Chunbao Wang\inst{3} \and
Deyu Meng\inst{2,4} \and
Chen Li\inst{1,2}}


\authorrunning{Z. Gao et al.}

%
\institute{School of Computer Science and Technology, Xi'an Jiaotong University, Xi'an, Shaanxi 710049, China \and
 National Engineering Lab for Big Data Analytics, Xi'an Jiaotong University, Xi'an, Shaanxi 710049, China
\and Department of Pathology, the First Affiliated Hospital of Xi’an Jiaotong University, Xi’an, 710061, China \and School of Mathematics and Statistics, Xi'an Jiaotong University, Xi'an, Shaanxi 710049, China\\
\email{gzy4119105156@stu.xjtu.edu.cn}}


%
\titlerunning{Instance-based ViT for Subtyping of pRCC in Histopathological Image}
\maketitle              
\begin{abstract}


Histological subtype of papillary (p) renal cell carcinoma (RCC), type 1 vs. type 2, is an essential prognostic factor. 
The two subtypes of pRCC have a similar pattern, i.e., the papillary architecture, yet some subtle differences, including cellular and cell-layer level patterns. 
However, the cellular and cell-layer level patterns almost cannot be captured by existing CNN-based models in large-size histopathological images, which brings obstacles to directly applying these models to such a fine-grained classification task. 
This paper proposes a novel instance-based Vision Transformer (i-ViT) to learn robust representations of histopathological images for the pRCC subtyping task by extracting finer features from instance patches (by cropping around segmented nuclei and assigning predicted grades). 
The proposed i-ViT takes top-K instances as input and aggregates them for capturing both the cellular and cell-layer level patterns by a position-embedding layer, a grade-embedding layer, and a multi-head multi-layer self-attention module. 
To evaluate the performance of the proposed framework, experienced pathologists are invited to selected 1162 regions of interest from 171 whole slide images of type 1 and type 2 pRCC. 
Experimental results show that the proposed method achieves better performance than existing CNN-based models with a significant margin.

\end{abstract}
\section{Introduction}

\begin{figure}[t]
\centering
\subfigure[]{
\includegraphics[width=0.4\textwidth]{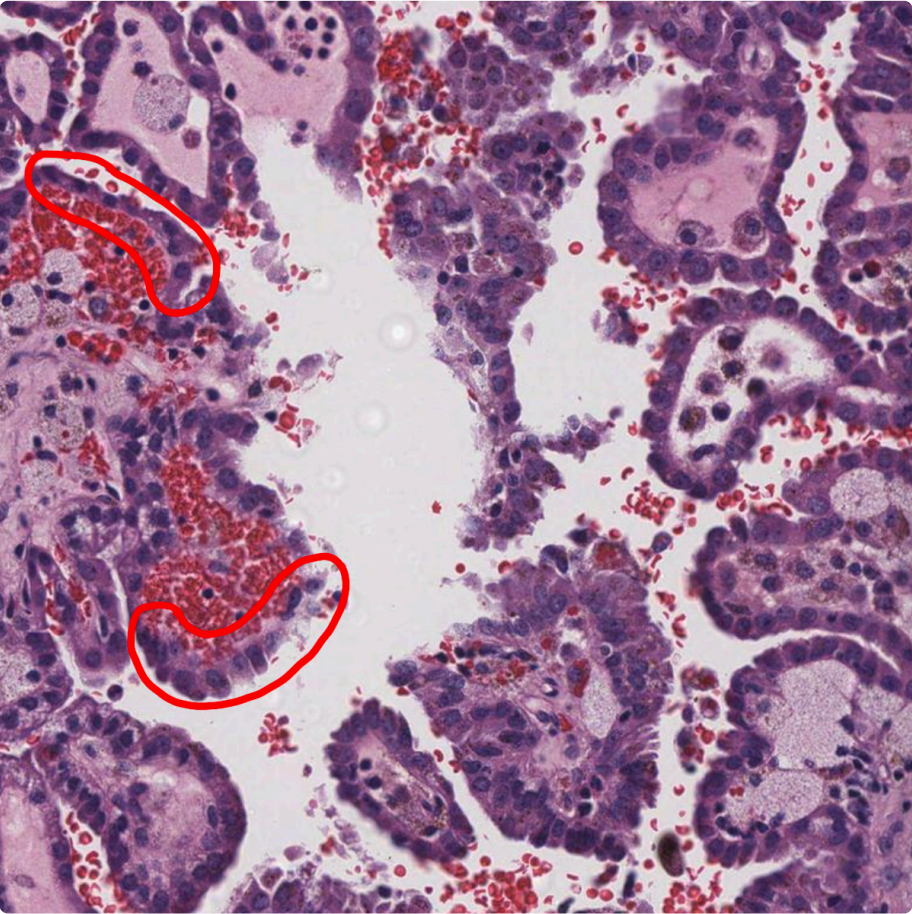}
}
\subfigure[]{
\includegraphics[width=0.4\textwidth]{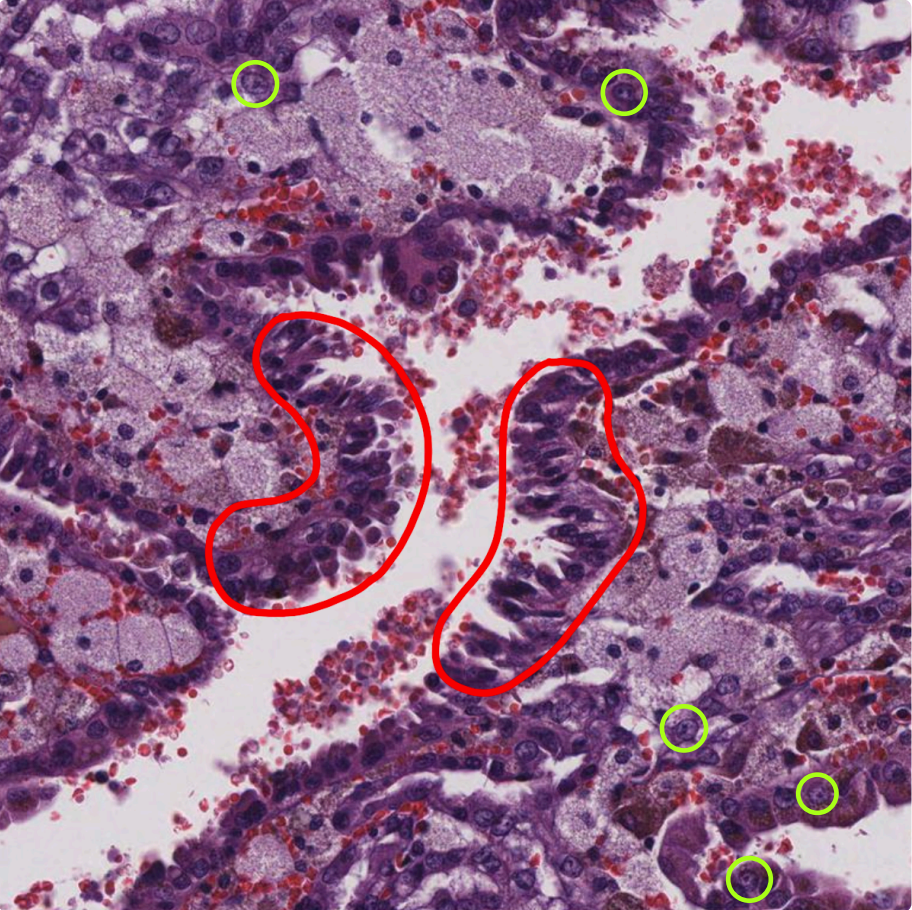}
}
\caption{
An illustration of papillary renal cell carcinoma (a) type 1 and (b) type 2. 
Red circles highlight the different papillary structures: (a) a single layer, and (b) pseudostratified cell layers. 
Green circles in (b) indicate some high-grade nuclei that exclusively exist in type 2. 
Note that all the areas marked by green and red circles are relatively small and randomly distributed.
} 
\label{fig1}
\end{figure}

Papillary renal cell carcinoma (pRCC) is the second common type of RCC, which accounts for 10\% to 20\% of all RCC cases \cite{pRCCintro}. 
The International Society of Urological Pathology (ISUP) system \cite{isup} and other researches \cite{WONG2019721,leroy2002morphologic,pan2020effect} indicate that pRCC subtyping can provide valuable prognostic information.
For example, according to different subtypes, different treatment strategies that aim to improve the patients' survival rate can be precisely conducted.
Based on the histologic features, pRCC can be classified into two subtypes: type 1 and type 2 \cite{WHO}.
Type 1 composes of papillae covered with a single layer of small cells with low ISUP grade containing basophilic cytoplasm, while type 2 tumors have pseudostratified cells with high ISUP grade containing eosinophilic cytoplasm. 

As a fundamental task in digital pathology analysis, cancer subtyping has attracted much attention from researchers in the computer vision domain.
Recently, several convolutional neural network (CNN)-based models have been proposed for different cancer of various organs, \textit{e.g.}, breast \cite{couture2018image}, lung \cite{coudray2018classification} and kidney \cite{gao2020renal}.
The key point of these models is to identify the subtypes of these cancers according to certain histological features that can be easily extracted by CNN, such as architectural patterns of the tumors. 
However, pRCC subtyping is a fine-grained classification task because type 1 and type 2 pRCC have similar architectural patterns (i.e., papillary structure), which bring obstacle to apply existing models directly to the pRCC subtyping task.

In the pRCC pathology analysis routine, pathologists distinguish type 1 and type 2 pRCC mainly based on two fine features: the \textit{cellular level} and the \textit{cell-layer level} patterns (see Fig.\ref{fig1} for detail). 
Unfortunately, it is almost impossible for traditional CNN-based models to capture these two essential characteristics.
Some works attempt to construct features manually based on the predicted results of nuclei segmentation and classification models for extracting \textit{cellular level} patterns \cite{Wang2056,Chenge91}.
However, these hand-craft features are task-specific that hard to generalize to different digital pathology tasks.
In the term of \textit{cell-layer level} patterns, existing CNN-based models cannot percept the relative position patterns between cells (which with papillary structures) and layers composed by their around clustering cells. 
Moreover, both \textit{cellular level} and \textit{cell-layer level} patterns are relatively small and randomly distributed in different locations of a pRCC histopathological image, thus making it challenging to highlight such patterns in the extracted features of a traditional CNN.

To tackle the aforementioned challenges, in this paper, we propose an novel framework based on Transformer \cite{NIPS2017_3f5ee243}, namely instance-based Vision Transformer (i-ViT), for the pRCC subtyping. 
The central idea of the proposed i-ViT is to capture instances features first by extracting instance-level patches that each patch includes a nucleus with part of the surrounding background.
Then aggregate selected instance-level features by the self-attention mechanism for further capturing \textit{cellular level} and \textit{cell-layer level} features, and encode both of them into the final image-level representation. 
Specifically, instance-level patches are extracted and classified by a segmentation network with additional classification channels to acquire graded nuclei segments.  
Then we select the top-K instance-level patches according to nuclei grade and size. 
After that, we use a CNN to embed each selected instance-level patch into instance-embedding.
Considering the importance of the relative position between nuclei and grade information of nuclei, we encode the nuclei position and grade into the instance-embedding through a position embedding layer and a grade embedding layer, respectively.
Finally, we use multi-head multi-layer self-attention to integrate the instance-embeddings with position and grade label encodings.

The main difference between our i-ViT and existing vision Transformer models, such as vision Transformer (ViT) \cite{dosovitskiy2020image}, is that our proposed i-ViT only uses instance-level patches, which can learn key features and reduce the computational complexity, while ViT takes all split image patches into account.
Meanwhile, we propose to leverage the nuclei grade features by applying the nuclei label embedding. 
Extensive experiments are conducted on a dataset derived initially from the kidney renal papillary cell carcinoma project of The Cancer Genome Atlas (TCGA-KIRP) database. 
The experimental results demonstrate that i-ViT achieves better performance than other comparison methods.

\section{Method}

\begin{figure}[t]
\centering
\includegraphics[width=\textwidth]{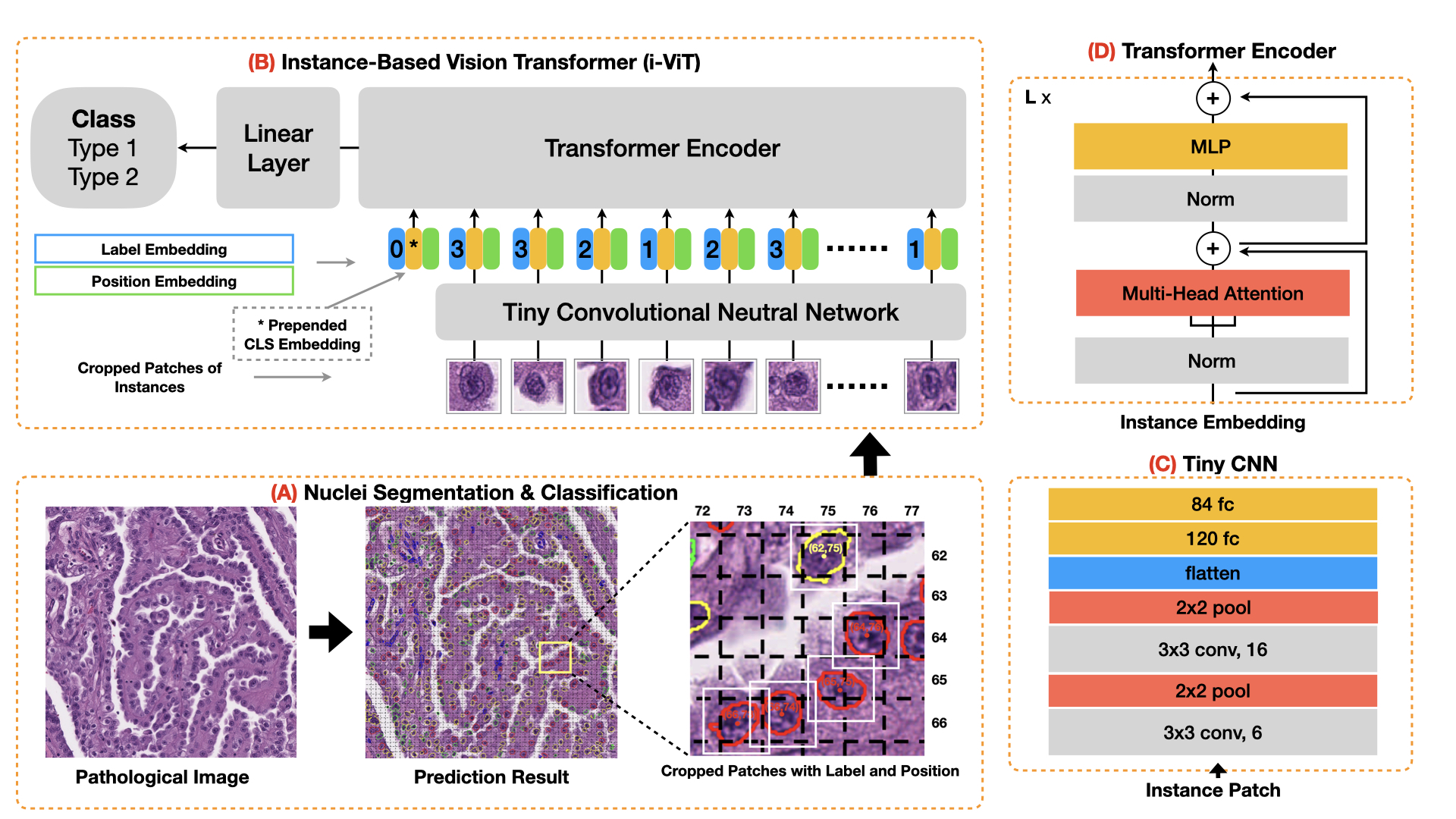}
\caption{
The proposed framework for papillary renal cell carcinoma subtyping. 
It include two stages: (A) for obtaining instances features, (B) for pRCC subtyping via extracting and aggregating instance-level features.
} 
\label{fig2}
\end{figure}

An overview of our proposed framework for pRCC subtyping is depicted in Fig. \ref{fig2}. 
The pipeline can be briefly described as follows: 
(1) Training a nuclei segmentation and classification model for segmenting nuclei and predicting their grade in an input histopathological image. 
(2) Extracting patches with assigned grades as instances according to the predicted results of step (1). 
(3) Select top-K instances and input them into the proposed i-VIT for predicting the subtype of the input image. 
The proposed i-ViT is illustrated in Fig. \ref{fig2}(B), and it is composed of a tiny CNN and a Transformer encoder.
The corresponding architectures of the tiny CNN and the Transformer encoder are shown in Fig. \ref{fig2}(C) and (D).

\subsection{Nuclei Segmentation and Classification}
To capture the fine features of pRCC subtypes, we need to segment out nuclei from the original image to avoid redundant information in a large-scale histopathological image and highlight the critical information for the downstream subtyping network.
Considering that nuclei grades are an essential basis for subtyping, we predict each segmented nucleus's grade as well. 
To this end, we train a network for jointly learning the nuclei segmentation and the classification tasks. 
We have evaluated several deep learning models, such as U-Net, HoVer-Net \cite{Hovernet}, and Micro-Net \cite{micro-net}. 
As Micro-Net achieves better comprehensive performances on accuracy and inference speed than other models, we adopt it for the segmentation and classification task in our framework.

\subsection{Instance-Based Vision Transformer}

After training the nuclei segmentation and classification model, we crop instance-level patches from the original image according to the segmentation results as shown in Fig. \ref{fig2} (A).
Each patch is in a size of $P \times P, P=64$ and centered by a nucleus point, in which a whole nucleus and part of the surrounding background are included. 
Therefore, the cropped patches carry the cellular and location (whether the cells locate on the papillary structure or not) features.

The standard Transformer \cite{NIPS2017_3f5ee243} is restrict to only take the sequences of 1D embeddings as input by its architecture.
To extract 1D feature embeddings $S\in \mathbb{R}^{N\times D}$ from a sequence of 2D instance-level patches $X\in \mathbb{R}^{N\times P\times P\times C}$ (Eq. \ref{eq1}), we construct a tiny CNN model which consists of two convolution and two fully-connection layers as the trainable patch embedding projection $\pounds$. 
$D$ denotes the patch embedding dimension and serves as the latent vector size for all the Transformer layers, $C$ is the number of input channels. 
Note that, different images contain different number of nuclei, similar to the standard processing approach of NLP, we set a hyper-parameter $N$ which is the number of instances, also as the input sequence length of Transformer to handle this variation. 
Due to the importance of nucleus grade and size for subtyping, we remove the endothelial nuclei and sort other nuclei instances (tumor nuclei) by grade and size, which are easily obtained from nuclei segmentation and classification results, then the top $N$ of nuclei instances are selected as the input of tiny CNN.

Following the similar idea of ViT \cite{dosovitskiy2020image}, we assign a learnable class embedding $s_{cls}^{0}\in \mathbb{R}^{1\times D}$ to learn the image representation that can be regarded as an aggregation of all the selected instance embeddings. 
During the training and testing process, a single linear layer classification head $s_{cls}^{L}$ is attached (Eq. \ref{eq4}). 
$L$ is the number of Transformer layer.

The nuclei positional information, which is the coordinates of each nucleus in the image, reflects the relative position between nuclei and is related with the \textit{cell-layer level} patterns. 
Since the histopathological image size is large and meaningless to learn the redundant position embedding, we grid the image coordinates into 1/20 of the original size and use the grid's position where the center of a nucleus is located to be the position of the nucleus, as shown in Fig .\ref{fig2}(A). 
Thereby, the position embeddings $s_{X} , s_{Y}\in \mathbb{R}^{(N+1)\times D/2}$ are learned from the nuclei position with two axes. 
Then we concatenate the X and Y embeddings to form the final position embeddings of the nuclei. 
To encode the nuclei grades information, we employ a grade embedding to integrate the ISUP nucleolar grade features as $s_{grade}\in \mathbb{R}^{(N+1)\times D}$. 
Finally, the position embeddings and grade embeddings are added to instance embeddings (Eq. \ref{eq1}).

The architecture of our Transformer encoder is similar to ViT. 
It consists of repeated stacked multi-head self-attention (MSA) and multilayer perceptron (MLP) blocks (Eq. \ref{eq2},\ref{eq3}). 
The MLP consists of two hidden layers with GELU activation. Layernorm (LN) and residual connection are applied in every block. 

\begin{equation}
S_{0}=\left [s_{cls}^{0};\pounds \left ( x_{1} \right );\pounds \left ( x_{2} \right );...;\pounds \left ( x_{N} \right ) \right ] + \left [ s_{X}, s_{Y} \right ] + s_{grade}
\label{eq1}
\end{equation}

\begin{equation}
\hat{S}_{l}=\textup{MSA}\left ( \textup{LN}\left ( S_{l-1} \right ) \right ) + S_{l-1} \qquad l=\left \{ 1...L \right \}
\label{eq2}
\end{equation}

\begin{equation}
S_{l}=\textup{MLP}\left ( \textup{LN}\left ( \hat{S}_{l} \right ) \right ) + \hat{S}_{l} \qquad l=\left \{ 1...L \right \}
\label{eq3}
\end{equation}

\begin{equation}
y=\textup{Linear}\left ( s_{cls}^{L}\right )
\label{eq4}
\end{equation}
where y is the output vector of i-ViT, it can be transformed as predicted probabilities via a softmax function.

\section{Experiment}

\subsection{Dataset}
For nuclei segmentation and classification, we adopt a nuclei grading dataset introduced in \cite{gao2021nuclei}, it contains 1000 region of interests (ROIs) that are selected from histopathological images of clear cell RCC (ccRCC) and pRCC.
All the tumor nuclei with grade 1, 2, 3, and endothelial nuclei are labeled in these ROIs.  
Although this dataset is originally proposed for ccRCC, the nuclei grading system (ISUP) used in the annotation is recommended and validated on both ccRCC and pRCC. 
We only use the images of pRCC from the testing set to evaluate the model performance. 

The pRCC subtyping dataset comprised 171 diagnostic whole slide images (WSI) from 171 patients (scanned at 40x), of which 62 of type 1, 109 of type 2. 
These WSIs are downloaded from KIRP project of TCGA and re-diagnosed by two experienced pathologists. 
The pathologists have selected a total of 1162 (613 vs. 549) ROIs in $2000 \times 2000$ size, approximately 10 ROIs for every type 1 case, 5 for type 2.  
We randomly divide the dataset into three subsets in patient-level, training (60\%), validation (20\%), and testing (20\%) set. 
This dataset and source code of our work are available at:
\url{https://dataset.chenli.group/home/prcc-subtyping} and \url{https://github.com/ZeyuGaoAi/Instance_based_Vision_Transformer}.

\subsection{Implementation Details}

\subsubsection{nuclei segmentation and classification}
To avoid bias from different staining conditions, we perform the staining normalization and image augmentations (flip, rotate, blur) on all images. 
For the training process, the batch size is 8, the learning rate is 1.0e-4, and decreases to 1.0e-5 after 25 epochs. 
Adam optimizer is adopted, and the total training epochs 50. 
The weighted panoptic quality (wPQ) used in the MoNuSAC challenge \cite{monusac2020} is adopted for evaluation in which the weight of each class is 1. 
The wPQ of Micro-Net is up to 0.5, which is an acceptable performance for pRCC nuclei segmentation and classification.

\subsubsection{Subtyping of pRCC}
Four methods are adopted for performance comparison. 
1. A CNN model (ResNet-34) without considering nuclei features, the model's input are the original ROIs, denoted by CNN-ORI. 
2. The traditional classifier (Decision Tree) with the grade distribution of each ROI, denoted by DT-G. 
3. The gradient boosting decision tree (GBDT) with the grade distribution and hand-craft cellular features aggregation, denoted by GBDT-GH. 
4. The single Transformer encoder with hand-craft cellular features, denoted by i-ViT-H.

The grade distribution is obtained from the nuclei classification result of each ROI.
We normalize the number of nuclei in each grade (1,2 and 3) to get the corresponding rate for DT-G. 
For hand-craft cellular feature extraction, four types of features, which are size, shape, color, and distance to neighbors, are extracted for each nucleus. 
We select ten cellular features: the area of nucleus, the major and minor axes lengths of a nucleus and the ratio between these two lengths, the mean pixel values in three color channels (RGB) of a nucleus, three distances (maximum, minimum, mean) to neighbors in the Delaunay triangulation. 
Similar to \cite{Chenge91}, we adopt five distribution statistics, i.e., mean, std, skewness, kurtosis, entropy, and a 10-bin histogram to integrate the selected cellular features into one image-level feature.  
Finally, 150-dimensional hand-craft features and three grade-distribution features are generated for the third method GBDT-GH. 
The fourth model i-ViT-H only takes the ten cellular features of each nucleus as the input of the Transformer encoder. 
The grade embedding and the prepend class token are also adopted.

We implement all the methods in experiments with the open-source library Pytorch 1.6 and scikit-learn 0.24.0 on a work-station with four NVIDIA 2080Ti GPUs. 
For traditional classifiers (DT, GBDT), we use the default parameters from scikit-learn. 
Except for the image-net pre-trained ResNet-34 is used in the first comparison method, we do not use any other pre-trained models. 
For deep models, the adam optimizer is adopted for training, and the maximum epochs are 50.  
The learning rate is 1.0e-3 initially and decreases to 1.0e-4 after 30 epochs for CNN-ORI and i-ViT, and is 1.0e-2 (1.0e-3 after 30 epochs) for i-ViT-H. 
Due to the low-dimensional embeddings of i-ViT-H, the number of heads, layers, and hidden dimensions are set to 2, 1, and 32, respectively.  
For i-ViT, the number of heads, layers, and hidden dimensions are set to 12, 12, and 128, respectively.  
The input sequence length $N$ is set to 500 for i-ViT-H and i-ViT. 
A linear warm-up learning rate schedule is used in i-ViT and i-ViT-H for 10 epochs.

\subsection{Results}

\begin{table}[]
\caption{The comparison results, overall-accuracy (Acc), precision (Prec), recall (Rec), F1-score (F1) for pRCC subtyping.}
\label{tab1}
\centering
\begin{tabular}{|c|c|c|c|c|c|c|c|}
\hline
\multirow{2}{*}{Model} & All    & \multicolumn{3}{c|}{Type 1} & \multicolumn{3}{c|}{Type 2} \\ \cline{2-8} 
                       & Acc\%   & Prec\%   & Rec\%   & F1\%   & Prec\%   & Rec\%   & F1\%   \\ \hline
CNN-ORI                & 75.11   & 77.39    & 74.17   & 75.74  & 72.81    & 76.15   & 74.44  \\ \hline
DT-G                   & 80.35   & 85.05    & 75.83   & 80.18  & 76.23    & 85.32   & 80.52  \\ \hline
GBDT-GH                & 88.21 & \textbf{94.29} & 82.50  & 88.00  & 83.06  & \textbf{94.50} & 88.41  \\ \hline
i-ViT-H                & 89.18   & 91.30    & 87.50   & 89.36  & 86.84    & 90.83   & 88.79  \\ \hline
i-ViT                  & \textbf{93.01} & 90.00 & \textbf{97.50} & \textbf{93.60}  & \textbf{96.97}  & 88.07 & \textbf{92.31}  \\ \hline
\end{tabular}
\end{table}

The overall accuracy, precision, recall, and F1-scores of each class are applied to evaluate the performance of subtyping models, and the classification results are shown in Table. \ref{tab1}. 
As we expected, CNN-ORI shows the worst performance because traditional convolution structures can hardly capture the small and randomly distributed features. 
GBDT-GH outperforms DT-G by about 8\% in ACC, which benefits from the rich hand-craft cellular features.
The i-ViT-H achieves better performance than GBDT-GH, where the improvements are due to the self-attention-based architecture. 
Moreover, our proposed i-ViT significantly outperforms other methods, which proves that the instance level features (nucleus patch, position, and grade embeddings) and the learning schema (learning with instance) we designed are effective for this fine-grained classification task.

\subsubsection{Sensitivity analysis}

\begin{figure}[t]
\centering
\subfigure[]{
\includegraphics[width=0.43\textwidth]{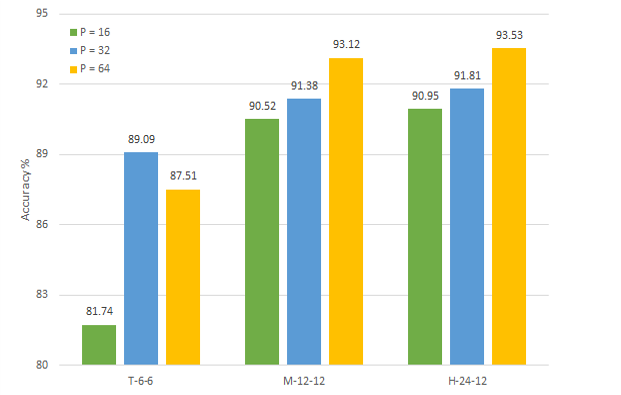}
}
\subfigure[]{
\includegraphics[width=0.43\textwidth]{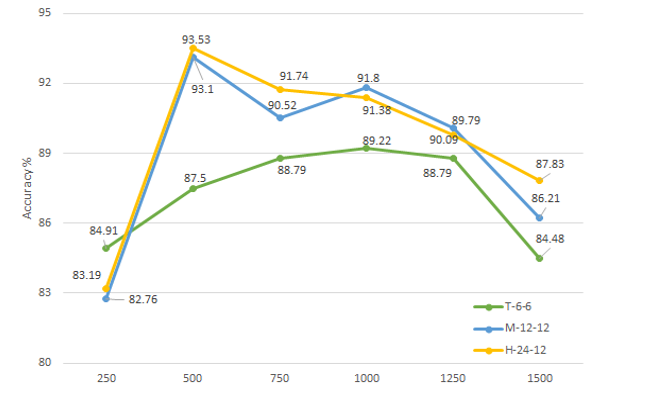}
}
\caption{
The i-ViT performance under various parameters. 
The T-6-6, M-12-12, and H-24-12 indicate three Transformers (tiny, middle, and huge), the first and second digits are the number of layers and heads, and all hidden dimensions are 128. 
(a) Different sizes of instance patches with the input sequence length $N=500$. 
(b) Different input sequence lengths with the size of instance patches $P=64$.}
\label{fig3}
\end{figure}

Because our proposed subtyping framework relies on parameters of the i-ViT model, three main parameters are analyzed:
1) The size of instance patches $P$ for tiny CNN, $P$=16, 32, and 64. 
2) The input sequence length $N$ of the Transformer encoder, from 250 to 1500 at an interval of 250. 
3) Three Transformer encoders with different scales (i.e., T-6-6, M-12-12, and H-24-12).

From Fig. \ref{fig3}, we can observe that the performance of the i-ViT is related to these three parameters. 
The performances of the huge and middle scale models are competent in most of the parameter settings, and the tiny scale model has the worst performance due to the under-fitting problem. 
The smallest size patches ($P=16$) lead to worse performance because these patches only contain part of nuclei, see Fig. \ref{fig3}(a). 
Also, in Fig \ref{fig3}(b), the performances are worse with $N=250$. 
We suppose that the models cannot extract discriminate features from limited instances, and these models are relatively stable with input sequence lengths (from 500 to 1000). 
Then model performances decrease rapidly from 1000 to 1500 because the number of tumor nuclei extracted from most of the images is less than 1250.  

\section{Conclusion}
Subtyping of pRCC has poor clinical consistency, the diagnostic difference between our pathologists and TCGA is more than 35\%, so it is necessary to design an automated subtyping model for pRCC. 
In digital pathology, this task is a fine-grained image classification task based on some detailed features and hard to solve by traditional CNN models. 
In this paper, we introduced a new learning schema, "learning with instances," for this particular task. 
Unlike the standard ViT that considers every pixel of each image, we integrated the instance information extracted from the nuclei segmentation and classification results with the ViT to form the i-ViT. 
The i-ViT only takes pixels of each instance into account, making the model ignore useless background information and pay more attention to learning useful features.

\subsubsection{Acknowledgements.}
This work has been supported by National Natural Science Foundation of China (61772409); This work has been supported by the National Key Research and Development Program of China (2018YFC0910404); The consulting research project of the Chinese Academy of Engineering (The Online and Offline Mixed Educational Service System for “The Belt and Road” Training in MOOC China); Project of China Knowledge Centre for Engineering Science and Technology; The innovation team from the Ministry of Education (IRT\_17R86); and the Innovative Research Group of the National Natural Science Foundation of China (61721002).
The results shown here are in whole or part based upon data generated by the TCGA Research Network: https://www.cancer.gov/tcga.

%
%
%
\bibliographystyle{splncs04}
\bibliography{refer}

\begin{thebibliography}{10}
\providecommand{\url}[1]{\texttt{#1}}
\providecommand{\urlprefix}{URL }
\providecommand{\doi}[1]{https://doi.org/#1}

\bibitem{pRCCintro}
Incidence and long-term prognosis of papillary compared to clear cell renal
  cell carcinoma – a multicentre study. European Journal of Cancer
  \textbf{48}(15),  2347--2352 (2012). \doi{10.1016/j.ejca.2012.05.002}

\bibitem{isup}
Delahunt, B., Cheville, J.C., Martignoni, G., Humphrey, P.A., Magi-Galluzzi,
  C., McKenney, J., Egevad, L., Algaba, F., Moch, H., Grignon, D.J., et~al.:
  The international society of urological pathology (isup) grading system for
  renal cell carcinoma and other prognostic parameters. The American journal of
  surgical pathology  \textbf{37}(10),  1490--1504 (2013)

\bibitem{WONG2019721}
Wong, E.C., {Di Lena}, R., Breau, R.H., Pouliot, F., Finelli, A., Lavallée,
  L.T., So, A., Tanguay, S., Fairey, A., Rendon, R., Richard, P.O., Lattouf,
  J.B., Kawakami, J., Mallick, R., Kapoor, A.: Morphologic subtyping as a
  prognostic predictor for survival in papillary renal cell carcinoma: Type 1
  vs. type 2. Urologic Oncology: Seminars and Original Investigations
  \textbf{37}(10),  721--726 (2019). \doi{10.1016/j.urolonc.2019.05.009}

\bibitem{leroy2002morphologic}
Leroy, X., Zini, L., Leteurtre, E., Zerimech, F., Porchet, N., Aubert, J.P.,
  Gosselin, B., Copin, M.C.: Morphologic subtyping of papillary renal cell
  carcinoma: correlation with prognosis and differential expression of muc1
  between the two subtypes. Modern pathology  \textbf{15}(11),  1126--1130
  (2002)

\bibitem{pan2020effect}
Pan, H., Ye, L., Zhu, Q., Yang, Z., Hu, M.: The effect of the papillary renal
  cell carcinoma subtype on oncological outcomes. Scientific Reports
  \textbf{10}(1), ~1--7 (2020). \doi{10.1038/s41598-020-78174-9}

\bibitem{WHO}
Moch, H., Cubilla, A.L., Humphrey, P.A., Reuter, V.E., Ulbright, T.M.: The 2016
  who classification of tumours of the urinary system and male genital
  organs—part a: Renal, penile, and testicular tumours. European Urology
  \textbf{70}(1),  93--105 (2016). \doi{10.1016/j.eururo.2016.02.029}

\bibitem{couture2018image}
Couture, H.D., Williams, L.A., Geradts, J., Nyante, S.J., Butler, E.N., Marron,
  J., Perou, C.M., Troester, M.A., Niethammer, M.: Image analysis with deep
  learning to predict breast cancer grade, er status, histologic subtype, and
  intrinsic subtype. NPJ breast cancer  \textbf{4}(1), ~1--8 (2018).
  \doi{10.1038/s41523-018-0079-1}

\bibitem{coudray2018classification}
Coudray, N., Ocampo, P.S., Sakellaropoulos, T., Narula, N., Snuderl, M.,
  Feny{\"o}, D., Moreira, A.L., Razavian, N., Tsirigos, A.: Classification and
  mutation prediction from non--small cell lung cancer histopathology images
  using deep learning. Nature medicine  \textbf{24}(10),  1559--1567 (2018).
  \doi{10.1038/s41591-018-0177-5}

\bibitem{gao2020renal}
Gao, Z., Puttapirat, P., Shi, J., Li, C.: Renal cell carcinoma detection and
  subtyping with minimal point-based annotation in whole-slide images. In:
  Medical Image Computing and Computer Assisted Intervention -- MICCAI 2020.
  pp. 439--448. Springer International Publishing, Cham (2020)

\bibitem{Wang2056}
Wang, S., Rong, R., Yang, D.M., Fujimoto, J., Yan, S., Cai, L., Yang, L., Luo,
  D., Behrens, C., Parra, E.R., Yao, B., Xu, L., Wang, T., Zhan, X., Wistuba,
  I.I., Minna, J., Xie, Y., Xiao, G.: Computational staining of pathology
  images to study the tumor microenvironment in lung cancer. Cancer Research
  \textbf{80}(10),  2056--2066 (2020). \doi{10.1158/0008-5472.CAN-19-1629}

\bibitem{Chenge91}
Cheng, J., Zhang, J., Han, Y., Wang, X., Ye, X., Meng, Y., Parwani, A., Han,
  Z., Feng, Q., Huang, K.: Integrative analysis of histopathological images and
  genomic data predicts clear cell renal cell carcinoma prognosis. Cancer
  Research  \textbf{77}(21),  e91--e100 (2017).
  \doi{10.1158/0008-5472.CAN-17-0313}

\bibitem{NIPS2017_3f5ee243}
Vaswani, A., Shazeer, N., Parmar, N., Uszkoreit, J., Jones, L., Gomez, A.N.,
  Kaiser, L.u., Polosukhin, I.: Attention is all you need. In: Guyon, I.,
  Luxburg, U.V., Bengio, S., Wallach, H., Fergus, R., Vishwanathan, S.,
  Garnett, R. (eds.) Advances in Neural Information Processing Systems.
  vol.~30. Curran Associates, Inc. (2017)

\bibitem{dosovitskiy2020image}
Dosovitskiy, A., Beyer, L., Kolesnikov, A., Weissenborn, D., Zhai, X.,
  Unterthiner, T., Dehghani, M., Minderer, M., Heigold, G., Gelly, S.,
  Uszkoreit, J., Houlsby, N.: An image is worth 16x16 words: Transformers for
  image recognition at scale (2020)

\bibitem{Hovernet}
Graham, S., Vu, Q.D., Raza, S.E.A., Azam, A., Tsang, Y.W., Kwak, J.T., Rajpoot,
  N.: Hover-net: Simultaneous segmentation and classification of nuclei in
  multi-tissue histology images. Medical Image Analysis  \textbf{58},  101563
  (2019). \doi{10.1016/j.media.2019.101563}

\bibitem{micro-net}
Raza, S.E.A., Cheung, L., Shaban, M., Graham, S., Epstein, D., Pelengaris, S.,
  Khan, M., Rajpoot, N.M.: Micro-net: A unified model for segmentation of
  various objects in microscopy images. Medical Image Analysis  \textbf{52},
  160 -- 173 (2019). \doi{10.1016/j.media.2018.12.003}

\bibitem{gao2021nuclei}
Gao, Z., Shi, J., Zhang, X., Li, Y., Zhang, H., Wu, J., Wang, C., Meng, D., Li,
  C.: Nuclei grading of clear cell renal cell carcinoma in histopathological
  image by composite high-resolution network (2021)

\bibitem{monusac2020}
Verma, R., Kumar, N., Patil, A., Kurian, N., Rane, S., Sethi, A.: Multi-organ
  nuclei segmentation and classification challenge 2020  (02 2020).
  \doi{10.13140/RG.2.2.12290.02244/1}

\end{thebibliography}

\end{document}